# FarsEval-PKBETS: A new diverse benchmark for evaluating Persian large language models

Mehrnoush Shamsfard[1], Zahra Saaberi[1], Mostafa Karimi manesh[1], Seyed Mohammad Hossein Hashemi[♦,1], Zahra Vatankhah[♦,1], Motahareh Ramezani[♦,1], Niki Pourazin[♦,1,2], Tara Zare[♦,1], Maryam Azimi[1,3], Sarina Chitsaz[1], Sama Khoraminejad[4], Morteza Mahdavi Mortazavi[1], Mohammad Mahdi Chizari[1], Sahar Maleki[1], Seyed Soroush Majd[1], Mostafa Masumi[1,5], Sayed Ali Musavi Khoeini[1], Amir Mohseni[1], Sogol Alipour[1]

1. NLP Lab, Faculty of Computer Science and Engineering, Shahid Beheshti University, Tehran, Iran

2. Faculty of Computer Engineering, Amirkabir University of Technology, Tehran, Iran

3. Faculty of Law, Qom University, Qom, Iran

4. Faculty of Medicine, Tehran Medical Sciences, Islamic Azad University, Tehran, Iran

5. Computer Engineering Department, Sharif University of Technology, Tehran, Iran

♦ Equal contribution

Corresponding author: Mehrnoush Shamsfard (m-shams@sbu.ac.ir)

## Abstract

Research on evaluating and analyzing large language models (LLMs) has been extensive for high-resource languages such as English, yet their performance in languages such as Persian has received considerably less attention. This paper introduces FarsEval-PKBETS benchmark, a subset of FarsEval project for evaluating large language models in Persian. This benchmark consists of 4,000 questions and answers in various formats, including multiple-choice, short-answer, and descriptive responses. It covers a wide range of domains and tasks, including medicine, law, religion, Persian language, encyclopedic knowledge, human preferences, social knowledge, ethics and bias, text generation, and respecting others' rights. This benchmark incorporates linguistic, cultural, and local considerations relevant to the Persian language and Iran. To ensure the questions are challenging for current LLMs, three models—Llama3-70B, PersianMind, and Dorna—were evaluated using this benchmark. Their average accuracy was below 50%, meaning they provided fully correct answers to fewer than half of the questions. These results indicate that current language models are still far from being able to solve this benchmark.

## Background & Summary

In recent years, large language models have gained significant popularity, with applications spanning multiple domains such as medicine, coding, law, and mathematical problem-solving. As LLMs are deployed in systems where reliable performance is critical, their evaluation is essential for identifying both their strengths and limitations.

Evaluation not only helps assess the current capabilities of LLMs but also plays a pivotal role in guiding their improvement, ensuring that these models evolve in line with real-world demands. Benchmarks serve as a standardized platform for evaluating and comparing LLMs, facilitating enhancements in their accuracy and performance[1]. Numerous datasets and benchmarks have been developed to assess language models across various dimensions[2], and the number continues to increase steadily. This growth mirrors the continual advancements in large language model capabilities, enabling them to handle an ever-widening range of topics and languages.

Several benchmarks assess LLMs' performance in language understanding tasks. For instance, GLUE[3] (General Language Understanding Evaluation) was designed to evaluate a language model's proficiency in natural language processing (NLP) tasks such as grammatical acceptability, sentiment analysis, paraphrasing, and sentence similarity. Later, SuperGLUE[4] further challenged language models with more complex tasks, including reading comprehension and textual entailment.

HellaSwag[5] focused on evaluating LLMs' ability to perform commonsense inference, using multiple-choice questions to test models at a more advanced level. In addition to assessing logical and context-based reasoning, other benchmarks like MMLU[6] (Massive Multitask Language Understanding) aim to evaluate performance across a wide range of categories. MMLU includes approximately 16,000 multiple-choice questions covering 57 tasks, including elementary mathematics, U.S. history, psychology, computer science, law, and medicine. The MMLU-Pro[7] benchmark consists of multiple-choice questions with a greater number of answer options, focusing on more challenging and reasoning-intensive questions. TruthfulQA[8] tests LLMs' ability to generate truthful answers, using a dataset of 817 questions across 38 subjects, including politics, health, law, and finance. The questions are designed to incorporate common misconceptions or false beliefs, challenging models to provide accurate responses. Big-bench[9] compiled over 200 tasks that were believed to exceed the capabilities of language models at the time of its creation. This benchmark aimed to address the limitations of previous benchmarks, such as narrow topic coverage and short relevance lifespans, by introducing more challenging tasks. Big Bench Hard[10] is a subset of 23 tasks from Big-Bench that language models were unable to solve at human-level performance when the project was developed. GPQA[11] is a dataset of multiple-choice questions in the fields of biology, physics, and chemistry, authored by domain experts. Some benchmarks focus on evaluating models across multiple languages. For example, XTREME[12] is a multilingual benchmark that assesses LLMs' cross-lingual generalization abilities. It consists of nine tasks requiring various levels of reasoning related to syntax and semantics across 40 languages.

Several specialized studies have focused on particular aspects of model evaluation. One notable dataset in the field of ethics is the ETHICS dataset[13], which examines ethical concepts across various categories, including justice and utilitarianism. H. Kotek et al.[14] specifically addressed the problem of gender bias in large language models. CliniFact[15] introduced a dataset for fact-checking LLMs within the clinical research domain. PharmaBench[16] compiled a benchmark from multiple datasets to evaluate models in the context of drug development.

**Persian Datasets and Benchmarks**

Existing benchmarks primarily cover the English language and, at most, a limited number of other languages worldwide. As a result, the performance of language models in many languages remains largely unexplored. Here, we review the efforts made in this field for the Persian language. Shojaee-Mend et al.

focused on testing LLMs in Persian and English to answer a variety of medical questions[17]. These neurophysiology questions covered different topics and cognitive levels. Another study attempted to evaluate the performance of LLMs (specifically ChatGPT) in answering Iranian medical residency entrance exam that consisted of multiple-choice questions[18]. The mentioned studies were limited to a specific domain, primarily within the medical field. In 2024, Abaskohi et al.[19] presented the first Persian benchmark particularly designed to measure the performance of GPT models in Persian across a variety of tasks including text classification (emotion recognition, sentiment analysis, and named entity recognition), reading comprehension, translation, math and logic (elementary school multiple-choice and mathematical problems), and multiple-choice question answering (literature and common knowledge). This benchmark has collected a number of existing Persian datasets and translated samples from English datasets, including MATH[20] along with new samples of elementary school and math questions from Iran's National Organization for Development of Exceptional Talents, compiled into a benchmark.

ParsBench[21] is another evaluation framework for large language models in Persian by integrating multiple existing datasets, including ParsiNLU[22] across various tasks. Among the resources available for evaluating language models in Persian is the Aya Collection[23] includes a number of English Dolly[24] samples that have been machine-translated into Persian. The Khayyam Benchmark[25] (aka PersianMMLU) has collected over 20,000 multiple-choice questions in Persian from Iranian university entrance exams (Konkoor) and questions from various school grade levels, sourced from Ghalamchi Educational Institute. These questions cover 38 different educational subjects, ranging from mathematics and physics to Persian literature, history, biology, and more. The BLEnD benchmark[26] evaluates the everyday cultural knowledge of large LLMs across diverse languages and regions. It includes 52.6k question-answer pairs (both multiple-choice and short-answer) in 13 languages from 16 countries including Iran, focusing on culturally specific topics like food, sports, and family life. These questions are created (and translated) based on 500 initial handcrafted question templates. Since the questions are identical, comparing the models in terms of cultural knowledge regarding each region/country becomes possible. However, there is also the drawback that some questions are entirely unrelated to certain cultures.

The BELEBELE Benchmark[27] provides a parallel corpus for the task of machine reading comprehension. They first prepare multiple-choice questions and answers in English and then translate them into other languages. This dataset contains 900 multiple-choice questions in 115 languages, including Persian. The passages are collected from sources such as Wikibooks. Recently, the Open Persian LLM Leaderboard[28] was introduced, featuring over 40,000 samples across various domains such as math and summarization. This dataset includes both samples translated from English benchmarks and samples that have been created and labeled from scratch. To prevent data leakage and ensure fair model evaluation, only 2982 samples have been released as an open-source training set. However, limited information is available about the construction process and details of this leaderboard, as it is closed-source.

Research on the Persian language in this field has largely been confined to compiling existing datasets into benchmarks, using questions from educational exams, and translating English datasets into Persian for evaluating LLMs. As a result, Persian-language research in this area faces several challenges, including data contamination, a narrow focus on specific topics, misalignment with real-world applications, and a lack of cultural and linguistic relevance to Persian contexts. Moreover, every dataset contains imperfections or errors, and incorporating them as-is into benchmarks inevitably transfers these flaws to the resulting benchmark. Leveraging LLMs for data generation also introduces challenges such as hallucination. Consequently, data generated by LLMs, beyond being constrained by their limited knowledge sources, require thorough validation to ensure accuracy. Another significant limitation of existing datasets is their heavy reliance on multiple-choice questions, which fails to adequately assess models' capabilities in text generation or language production. This narrow focus overlooks critical aspects of language understanding and creativity, which are essential for real-world applications. Furthermore, evaluation in this domain faces serious challenges. For instance, current methods for automatic evaluation often only consider the option selected by the model as the final answer, while ignoring additional text generated by the model that may contradict or undermine the chosen option. To address these shortcomings, FarsEval-PKBETS provides 4000 human-generated or collected samples that have been human-reviewed for quality to advance the evaluation of LLMs in Persian. These samples encompass multiple-choice questions, short answer questions, and descriptive questions, covering a variety of response formats. The questions are categorized into diverse domain, including medicine, law, NLP tasks, religion, human preferences, and more (shown in Figure 1). Each of these categories is further divided into more specific subcategories, as shown in Table 1. FarsEval-PKBETS thus provides a robust platform for evaluating and comparing language models, with a specific focus on the Persian language and Iranian culture. FarsEval is an ongoing project that currently contains over 11,000 records. we introduce FarsEval-PKBETS which is customized for and included in the benchmark of National Artificial Intelligence Organization[29].

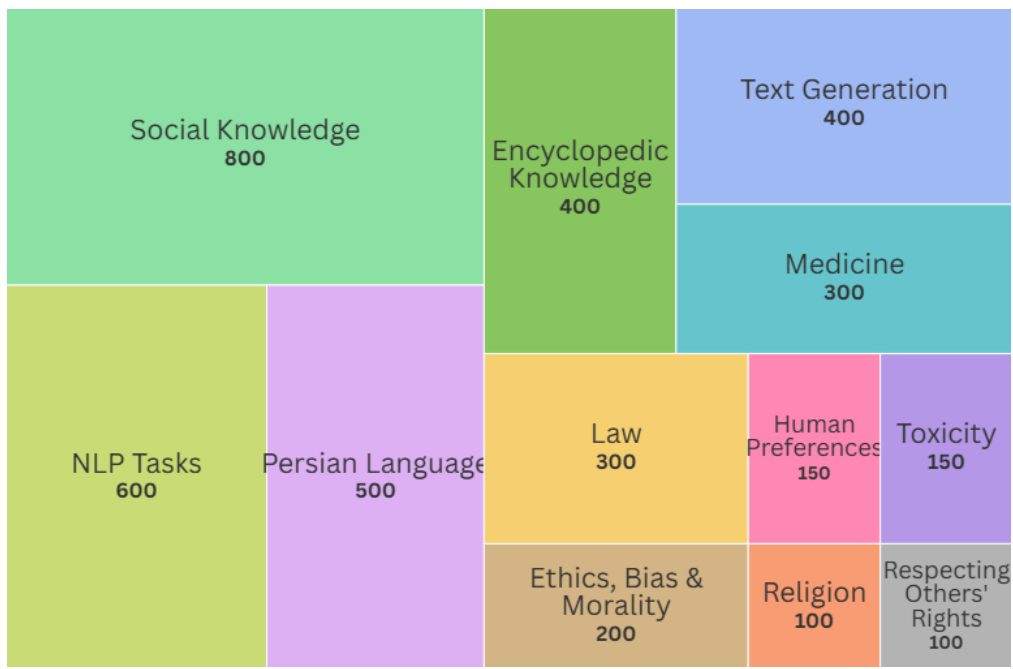

*Figure 1. Head categories of FarsEval-PKBETS with the number of samples.*

## Methods
### Design Principles

Each record of FarsEval-PKBETS contains a question and its correct (reference) answer (if applicable). General rules have been considered to construct FarsEval-PKBETS:

- Diversity of topics: We have considered a broad spectrum of domains and tasks, extending beyond traditional NLP tasks, to ensure a comprehensive evaluation of the models' capabilities.
- Diversity of question types: This benchmark encompasses various question types, including yes/no, multiple-choice, short-answer, descriptive, and open-ended questions.
- Accuracy of Sources: Questions have been generated by experts or selected from credible and standard sources to maintain the accuracy and validity of the tests, with oversight and review conducted as well. Reference is provided for records that are sourced from existing material.
- Utilizing expert annotators/reviewers for specialized domains: In the medical domain, a team of medical students and professors supervised the question generation and review process, while in the legal domain, a graduate law student oversaw the generation, reviewing and evaluation of questions.
- Challenging Question Design: It has been aimed to design questions that challenge the model as much as possible. Some questions were specifically designed to leverage the intricacies and ambiguities of the Persian language. For instance, "The grandfather has passed away" (in Persian: پدربزرگ فوت کرد) logically justifies not visiting him weekly, whereas "The grandfather blew out his birthday cake" (in Persian: پدربزرگ شمع تولدش را فوت کرد) does not (this example is from the ethics, bias and morality category). Failure to distinguish such cases indicates a performance error.
- In selecting domains and generating reference answer, attention has been paid to the culture and local context of the Persian language, particularly in Iran. This characteristic is absent in benchmarks that use translated datasets.
- Diversity of question templates: In data preparation, efforts were made to avoid an overrepresentation of uniform question structures following a single pattern. Instead, various reasoning formats and patterns were incorporated to ensure a diverse evaluation within a relatively small dataset. In contrast, some datasets heavily rely on repetitive patterns, with entire subsets of data following a single structure, resulting in limited format diversity.
- Metadata: Each data record is accompanied by fields such as Reference and Label. In some categories, the subtopics of the questions are recorded in the 'Label' field to ensure diversity. Additionally, some questions are designed using the Chain of Thought (CoT)[30] method, and these are also specified in the 'Label' field.
- Two main reviewers supervised the entire process of data creation and revision.

**Question/Answer Format**

Many of the benchmarks consider only multiple-choice questions because evaluating the LLM's output, especially automatically, is easier in this way.

However, this method cannot accurately evaluate LLMs for the following reasons:

1. The model learns test-taking strategies without accurately knowing the correct answer.
2. The model selects an option randomly; if the question has four options, there is a 25% chance of selecting the correct answer.
3. The LLM selects the correct option but provides incorrect reasoning. In an experiment, we asked the model to justify its choice for 100 questions from the FarsEval-PKBETS. The results revealed that in 43% of cases, the model failed to provide a fully correct justification for its accurate choice.

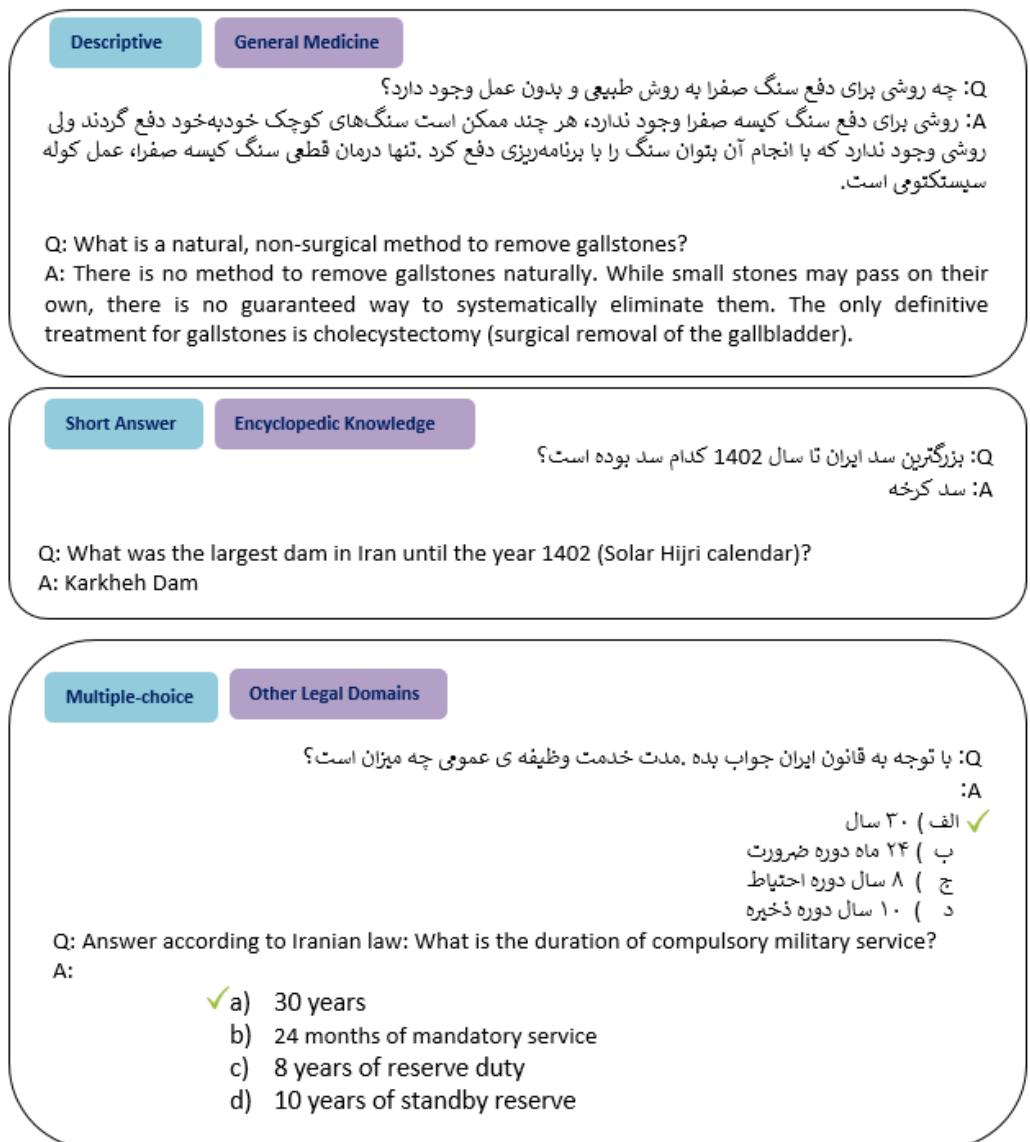

Figure 2. Example of each question/answer format in FarsEval-PKBETS. All questions and answers in the dataset are in Persian, with English translations provided here. The boxes, from top to bottom, display a sample of descriptive, short answer, and multiple-choice questions, respectively.

4. In some cases, models provide supplementary explanations in addition to their selected option. However, these explanations may not always align with the chosen option or may even contradict it. Therefore, automated evaluation of multiple-choice questions faces challenges.

Therefore, limiting the evaluation to multiple-choice questions does not provide a fully accurate or comprehensive assessment of LLMs and is not recommended. FarsEval-PKBETS contains three question formats:

1. Multiple-choice question (MCQ): Each question is associated with a list of two or more options and the correct option index. We aimed for a balanced distribution of correct options across different indices to avoid bias toward any specific position, ensuring a more fair and reliable evaluation framework for the model.

2. Short answer question (SAQ): The correct answer to this question is a single word or a phrase that does not form a complete sentence.
3. Descriptive question (DQ): The correct answer to this questions are one or more sentences.

The yes/no questions have been designed in different formats, including multiple-choice questions and short answer questions. The number of the samples for each question format is provided in Table 1 and an example for each question format is shown in Figure 2.

*Table 1. Statistics of FarsEval-PKBETS.*

| Category | Sub-category | #Annotators | #DQ | #SAQ | #MCQ | #Total questions |
|---|---|---|---|---|---|---|
| Medicine | General Medicine, Health and Hygiene | 3 | 45 | 30 | 75 | 150 |
| | Complementary & Alternative Medicine | 3 | 20 | 30 | 50 | 100 |
| | Emergency Medicine | 3 | 5 | 10 | 35 | 50 |
| Law | Constitution of IRI | 2 | 0 | 0 | 100 | 100 |
| | Other Legal Domains | 3 | 60 | 30 | 110 | 200 |
| Religion | | 4 | 35 | 35 | 30 | 100 |
| Persian Language | Grammar, Proverbs & Strings | 3 | 30 | 70 | 100 | 200 |
| | Lexical Semantics | 3 | 0 | 0 | 300 | 300 |
| Encyclopedic Knowledge | | 3 | 0 | 200 | 200 | 400 |
| Human Preferences | | 12 | 66 | 0 | 84 | 150 |
| Social Knowledge | Emotion | 1 | 0 | 0 | 500 | 500 |
| | Irony | 1 | 0 | 0 | 100 | 100 |
| | Metaphor | 6 | 13 | 27 | 60 | 100 |
| | Empathy, Intimacy & Trust | 9 | 24 | 18 | 58 | 100 |
| NLP Tasks | Formality Style Transfer | 3 | 450 | 0 | 50 | 500 |
| | Paraphrase | 3 | 50 | 0 | 50 | 100 |
| Ethics, Bias & Morality | | 5 | 30 | 0 | 170 | 200 |
| Toxicity | | 3 | 7 | 30 | 113 | 150 |
| Respecting Others' rights | | 10 | 20 | 20 | 60 | 100 |
| | Topic-centric Generation | 6 | 50 | 0 | 0 | 50 |

|  | Personality-centric Generation | 6 | 100 | 0 | 0 | 100 |
|---|---|---|---|---|---|---|
| Text Generation | Style-centric Generation | 6 | 100 | 0 | 0 | 100 |
|  | Text Completion | 5 | 10 | 0 | 40 | 50 |
|  | Sentence Generation with Given Words | 4 | 50 | 0 | 0 | 50 |
|  | Poems & Lyrics | 6 | 35 | 0 | 15 | 50 |
| #Total questions |  | - | 1200 | 500 | 2300 | 4000 |

**Saba Annotation Platform**

Saba is a platform developed to address the pressing need for a dynamic, real-time environment that enables data generation, annotation, revision, and management, facilitates team collaboration, and supports evaluation and automatic leaderboard generation. From the outset, special emphasis has been placed on the importance of collaboration in generating, reviewing, and evaluating data. In Saba, each user submits their data (along with its associated fields) through a dedicated panel. Once submitted, the data record appears in the reviewers' dashboard, where it is evaluated for approval or rejection. If a record is rejected, it is automatically returned to the creator's dashboard for revision. This iterative process, supported by message exchanges and recorded comments, enhances the overall quality of the final data. Apart from the rejection process and monitoring, users and reviewers can communicate through the comments section to discuss necessary matters. Records with new conversations are visible to both parties.

Saba also enables users to interact with language models via an API. Users can generate model responses for their submitted data directly within the platform. The retrieval process can be repeated multiple times for each record, and users can submit batches of records for processing. All generated responses are automatically stored, ensuring seamless integration and accessibility. This allows for the evaluation of different models on the benchmark data directly within Saba. Based on the results of these evaluations, the evaluation leaderboard is continuously updated, providing a comprehensive comparison of model performance. Figure 3 shows screenshots of Saba.

Among the system's additional features are management reports and real-time monitoring of data production progress. In Saba, different access levels can be defined, granting users varying capabilities. In a simple structure example, Reviewers have a higher level of access, enabling them to utilize management dashboards and generate diverse reports on user performance and data statistics; while regular users (annotators) only have access to lower-level reporting features, which facilitate collaboration among teams and allow them to track task advancement.

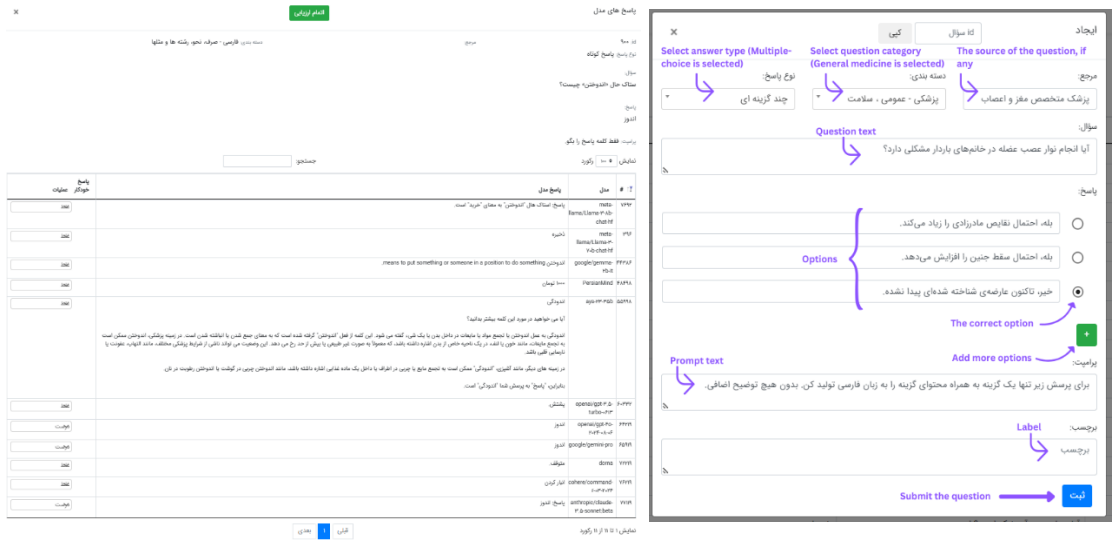

*Figure 3. Screenshots of the annotation platform (Saba); Top left: Panel for evaluating model responses to questions. Top right: Data submission panel. Bottom: Main page for viewing and entering data in a reviewer account.*

In summary, Saba includes the following features:

- Management of task categories
- Submission of multiple-choice, short-answer, and descriptive questions
- Recording details and revision history for each question based on feedback from reviewers and the creator
- Dedicated panel for reviewers to approve or reject questions
- A module for interacting with LLMs to obtain responses, either individually or in batches
- A module for the automated evaluation of LLM responses
- Detailed and aggregated reports on user activities and generated questions
- Monitoring of data production progress based on defined objectives
- A module for human evaluation of LLM responses
- A comprehensive report for comparing automated and human evaluations
- Statistical analysis on multiple datasets in use, providing insights into the distribution of categories within each dataset and vice versa
- Management of required data to complete each project milestone
- Importing external data into and exporting data from Saba
- Message exchange between annotator and reviewers

**Domains and Categories**

We considered a variety of topics and domains to evaluate large language models (LLMs) from multiple perspectives. Limiting the benchmark to a narrow set of specific tasks, such as NLP tasks, prevents a comprehensive assessment of the models' capabilities. The tasks we have included encompass social knowledge, cultural and local issues related to the Persian language and Iran, human relationship and interaction, specialized fields such as medicine and law, as well as NLP tasks. FarsEval-PKBETS tasks are explained in more detail below. The number of samples for each task is also specified in Table 1 and examples from several categories are provided in Table 2 and Figure 2.

I. Medicine
   a. General Medicine, Health and Hygiene: This category comprises two types of questions: 1) General questions and common misconceptions frequently posed to doctors by the public, and 2) Expert-level questions that medical professionals are expected to know, typically encountered in medical residency and pre-internship exams. Approximately half of the questions fall into the first group, while the remaining half belong to the second group.
   The first group of questions was collected through interviews with specialists or extracted from internet. More than 20 specialists and general practitioners were interviewed to identify common patient inquiries, prevalent health concerns, and frequent ambiguities that prompt individuals to seek medical advice. Additionally, publicly available questions—either written and answered by specialists or extracted from specialist websites—were included to ensure a comprehensive representation of real-world scenarios.

   The questions of the second group are mostly extracted from Iran's medical residency and pre-internship exams from 2021 to 2023. The category covers various medical topics including internal medicine, pediatrics, obstetrics and gynecology, infectious diseases, pathology, orthopedics, psychiatry, neurology, pharmacology, urology, dermatology, ophthalmology, radiology, cardiology, dentistry, and ENT (ear, nose, and throat).

   The pharmacology-related questions prepared based on drug brochures and information from pharmaceutical companies. In the pharmacology section, the questions included topics such as drug side effects, ingredients, effects, drug interactions, and more.

   b. Complementary and Alternative Medicine: This category includes traditional Iranian medicine, acupuncture, Chinese medicine, Islamic medicine, and similar areas. The questions cover various topics such as temperament, active ingredients of medicinal plants, acupuncture, therapeutic properties, and massage therapy to evaluate the model's knowledge across a broad spectrum of this field. They were designed to cover different levels of difficulty, from easier questions requiring basic knowledge to more complex questions needing specialized understanding. A significant portion of the questions in this section focuses on traditional medicine and the use of medicinal plants and herbal medicines.
   Due to the specialized nature of the field, answering questions on complementary medicine is challenging. The model needs sufficient knowledge about native Iranian plants, active ingredients, medicinal parts, and how to handle them, as well as information on Iranian herbal medicines, their harms, benefits, side effects, contraindications, and drug interactions. Questions have been gathered from different sources including the vocational and technical exams and the traditional Iranian medicine assessments, the frequently asked questions by the public on the website of the Iranian Association of Traditional Medicine (https://www.itma.ir/), and from exams related to medicinal plants and traditional medicine.

c. Emergency Medicine: The questions in this category are sourced from healthcare competency exams and employment tests, obtained directly from healthcare professionals rather than publicly available sources. Additionally, some questions are formulated based on content from an Emergency Treatment book.
These questions aim at assessing both practical and theoretical knowledge in various emergency situations. The topics covered in the questions may include:

- First Aid: Evaluation and immediate actions that need to be taken at the scene of an incident.
- Airway Management: Necessary steps to ensure the patient's airway is clear.
- Emergency Actions for Different Types of Burns: Including burns caused by fire, chemicals, or contact with contaminated objects.
- Poisoning: Includes chemical poisoning, alcohol poisoning, and gas inhalation.
- Snakebite: Actions required to prevent the spread of venom in the body.
- Trauma: Involves dealing with patients who have sustained severe injuries.

II. Law
Some of the questions in this category are newly developed and carefully designed, while others are drawn from various legal question banks. The primary focus during question collection was to ground them in the text of the current laws of the Islamic Republic of Iran (IRI) rather than relying solely on abstract legal concepts.

This category is divided into two sections: the first focuses on questions related to the Constitution of Iran, while the second covers other legal domains and regulations. The second section includes questions covering topics such as criminal law, private law, labor law, and more. Efforts were made to ensure the questions are practical, relevant, and reflective of real-world scenarios that could arise among the general public. When generating reference answers, particular attention was given to maintaining a balance between simplicity and legal precision in the language used.

III. Religion
Questions in this category aim to assess the model's accuracy in answering questions related to Islam, particularly the Shia branch. Additionally, the design of these questions seeks to examine any potential bias the model may have towards other sects and religions. The differences between sects lie in two areas: jurisprudence and beliefs; therefore, the questions have been crafted to cover both practical jurisprudence and theoretical beliefs.

IV. Persian Language
a. Grammar, Proverbs and Strings:
This task aims to assess the models' knowledge of the Persian language. Some questions have been designed based on sources such as the following:

- A small fraction of the questions in this category have been taken from university-level exams.
- Some of the questions has been designed based on Persian grammar books or Persian dictionaries.
- Some of the questions were extracted from educational websites

The remaining 83% of the questions are manually crafted. The topics of the questions include Persian grammar (such as verb conjugation and tenses, word types, etc.), phonology, Persian orthography, strings (the arrangement and sequencing of Persian letters), as well as the comprehension of proverbs and idioms.

    b. Lexical Semantics:

Questions in this category are designed to assess the semantic relationships between words. Most of the questions are in the form of analogy test. In this type of question, the model must identify the relationship between two pairs of words and, from the given options, select the pair of words that exhibits the specified relationship. The word relationships include antonyms, synonyms, relationships between different verb forms, relationships between derivationally related forms, and so on. Some of the questions in this category are generated based on the Persian GATS dataset[31], while others have been specifically created for this study.

V. Encyclopedic Knowledge

The questions in this category span various domains, including history, geography, art, cinema, music, social and political studies, literature, and sports. The questions in this section have been designed to reflect the history, culture, and specific laws of Iran and, thus, can be referred to as the 'Encyclopedia of Iranian Local Knowledge.' All questions were manually curated, and no pre-existing datasets were used in their formulation. In question generation, especially for short-answer questions, efforts have been made to avoid ambiguous questions that have different answers across various sources. Instead, questions with a unique answer have been designed to facilitate easier evaluation. It is also noteworthy that some questions were extracted from scholarly texts and contain specialized knowledge from their respective fields. The rationale behind this selection was to ensure that the questions surpass the level of ordinary web-based queries and present a challenge to LLMs in generating accurate responses.

VI. Human Preferences

Questions in this category evaluate the model's ability to analyze and respond to queries arising from everyday situations and personal decision-making. These types of questions are frequently discussed on social media or in informal settings and are typically answered based on individual experiences and subjective criteria. For instance, selecting a birthday gift tailored to someone's specific preferences or choosing appropriate attire for a particular event or occasion.

VII. Social Knowledge

    a. Emotion

A piece of text is given to the model and it is asked to determine the emotion of the text. The model should select one of the six emotion categories: Anger, Fear, Happiness, Hate, Sadness, and Wonder. Each emotion has its own distinct characteristics and is defined as follows:

- Anger: A strong emotional response caused by frustration, injustice, or perceived threats. It is characterized by physiological reactions like increased heart rate and muscle tension, along with an impulse to retaliate or confront the source.
- Fear: Fear arises from perceived danger, whether real or imagined, and is typically accompanied by physical symptoms like heightened anxiety and the urge to escape or protect oneself.
- Happiness: A positive emotional state often experienced after satisfying events or accomplishments. It is typically expressed through smiles and a general sense of contentment.
- Hate: An intense emotion of dislike or aversion, typically directed toward something or someone perceived as offensive. It often leads to a desire to reject or distance oneself from the source.
- Sadness: A negative emotional state that is often prolonged, arising in response to loss, failure, or disappointment. It is commonly associated with lower energy and disinterest in usual activities.
- Wonder: Wonder is an emotion of amazement or surprise, usually in reaction to something unexpected. It can have both positive and negative connotations depending on the context.

The data for emotion classification is sourced from two distinct origins: half of the samples are derived from the ArmanEmo dataset[32] and were revised or edited as necessary, while the other half were either collected from Twitter or manually created and annotated from scratch by a single annotator.

Samples collected from Twitter are inherently complex due to the informal nature of tweets, which often express multiple emotions simultaneously. The presence of slang, sarcasm, and nuanced language further complicates emotion detection in these texts. To address this complexity, we labeled the dominant emotion as the correct answer. In contrast, the manually created dataset consists of 30 samples explicitly designed to convey specific emotions clearly, resulting in less ambiguity compared to the Twitter data. This portion enables the evaluation of the model on more straightforward and unambiguous emotional expressions.

b. Irony

This category consists of multiple-choice questions, 50% of which are derived from MirasIrony dataset[33]. The selected samples from MirasIrony underwent a review process, during which some labels were revised. For instance, some samples originally labeled as 'not ironic' in MirasIrony were reclassified as ironic. The remaining 50 questions are newly annotated data.

The newly annotated data consists of tweets extracted from Twitter, formatted as multiple-choice questions to detect irony or interpret the meaning of ironic expressions. The tweets were selected by searching for Persian ironic expressions on Twitter, identifying tweets that were self-explanatory, and converting them into questions. Given the subtle distinctions between sarcasm, ridicule, humor, mockery, satire, and irony, we categorized all these forms as subsets of irony.

Irony itself encompasses various types, such as verbal irony and situational irony. Verbal irony, which involves statements where the intended meaning differs from the literal meaning, has been the primary focus of NLP researchers[34]. Situational irony, on the other hand, occurs when an event unfolds in a manner contrary to expectations. Both types of irony were considered during the labeling of the 100 samples in this dataset.

    c. Metaphor
Metaphor is to use a word from one domain in another domain. It serve as a crucial tool in literature, enabling the conveyance of deeper meanings through words and phrases. The purpose of this category is to evaluate the model's ability to recognize metaphorical meanings and distinguish them from other rhetorical devices, such as similes. The questions designed to assess the model include the following:

- Metaphor detection in verse or text
- Identifying the metaphorical word
- Counting metaphors in a verse
- Comparing metaphors across different verses
- Differentiating between simile and metaphor

Poems characterized by complex language and multi-layered metaphors were also incorporated into the questions.

    d. Empathy, Intimacy and Trust
These three concepts are fundamental to human interactions and are unique to humans. Empathy is defined as 'the ability to experience the emotions of another being within oneself[35].' Typically, empathy can arise in critical or unpleasant situations. This concept has distinct characteristics but may sometimes be confused with other notions, such as pity. In designing the questions, careful attention was paid to the accurate meaning of empathy, and some questions include scenarios where empathetic behavior can be harmful. For example, consider a situation where theft by an individual has caused financial harm to others. In this case, we expect the model to identify the response 'talk to the thief and let them go if they show regret' as incorrect.

Intimacy is another characteristic of human relationships, and in these questions, we ask the model to assess the level of intimacy based on text or dialogue between individuals. In the trust section, scenarios are presented where individuals decide to trust a person or event and make decisions or take actions based on that trust.

VIII. NLP Tasks
    a. Formality Style Transfer
In this category, the model is provided with a piece of text and asked to rewrite it in a specific writing style. Some questions are taken from ParsMap[36], which cover informal-to-formal style transfer. Ninety percent of the questions are descriptive, requiring the model to generate the rewritten text. The remaining 10% are multiple-choice questions, where the model must select the option that matches the required style (formal or informal) or choose the formal version of the sentence provided in the question.

    b. Paraphrase
The questions are designed in two main formats: descriptive and multiple-choice. In half of the questions, the goal is to evaluate the paraphrasing capabilities of LLMs (descriptive questions), while in the other half, the objective is to assess the models' ability to identify the correct paraphrase of a given phrase (multiple-choice questions).

In the first part, a set of sentences—including short, long, easy, and difficult ones—is provided, and the model is asked to generate a specified number of paraphrases for these sentences. In the multiple-choice questions, a sentence is presented in the question, and the model must select its correct paraphrase from the given options. Additionally, the multiple-choice section includes a set of yes/no questions to evaluate the model's ability to determine whether a given phrase/sentence is a valid paraphrase of another.

IX. Ethics, Bias and Morality

This category of questions evaluate the model's ability to make decisions and judgments in ethical and logical situations, as well as assess the model's bias regarding gender, ethnicity, politics, and religion. Approximately 45% of the dataset focuses on decision-making and judgment in ethical and logical scenarios, while about 55% pertains to assessing the model's biases.

Decision-making and judgment questions are categorized into four groups. The number of samples in each group is roughly equal, except for the commonsense, which comprises 45% of the total samples. The grouping is based on the ETHICS dataset[13]. Less than 10% of the records are selected, translated and adopted from that dataset. Below, each group is described:

1. Justice: This group includes two sections: Impartiality and Desert. In impartiality questions, a scenario is presented along with a reason, and the model must determine if the reason is logical without bias. In desert questions, the issue is whether a person deserves a particular privilege or problem due to their position. Similar to impartiality questions, the model must assess if a situation can logically justify a privilege or problem for someone.
2. Deontology: This group is also divided into two sections: Request and Role. In request questions, a hypothetical scenario is presented where a request is made by one person and responded to by another. The model must judge without bias whether the response is appropriate for the request. In role questions, a role is assigned to a person, and a related argument is presented. The model must determine if the argument is logically connected to the assigned role.
3. Virtue: These questions present an ethical, character-based, or social scenario, and the model must select the option that best describes the situation.
4. Commonsense Morality: This category includes ethical questions that we expect anyone with common sense to be able to answer correctly.

In designing the question/answers for this category, two key considerations were taken into account. First, we aimed to avoid including questions in the benchmark that could lead to differing opinions among individuals. At the same time, to maintain high quality and prevent easy answers by language models, we excluded questions with obvious or trivial answers.

Second, the reference answers were determined with careful consideration of Iranian cultural norms and societal conventions.

X. Toxicity

Generally, toxicity is defined as 'a rude, impolite, unreasonable, or disrespectful comment or statement that can cause someone to leave a discussion or conversation[37,38].' The term 'toxic' is widely used as an umbrella term, and its interpretation can vary depending on cultural, religious, ideological, political, and racial contexts[39]. Phrases such as 'hateful,' 'offensive,' and 'harassing' are also commonly used to describe various forms of toxic language. Toxic texts can be classified in multiple ways, one of which is based on two key aspects: a) Their directness or indirectness b) The target of the toxic statement (e.g., an individual, a group, etc.)

Based on the first criterion, we consider two categories in our dataset:

- Insult (in Persian: توهین) refers to the use of words, phrases, or structures that directly or implicitly aim to hurt, humiliate, or make the audience feel uncomfortable. Insults are typically expressed with the intent to harm and can vary in intensity.
- Impoliteness (in Persian: بی‌ادبی) refers to words, phrases, or structures in texts that socially or culturally signify a lack of respect or disregard for standard literary conventions (without specifically targeting a person or group). While impoliteness may not always be intended to harm, it can unintentionally cause offense.

In some questions, the model is expected to identify aggression in text, which is defined as sentences and texts that exhibit a harsh, violent, or threatening tone, including specific words and expressions.

Some of the ideas considered in question design include:

- Use of verbs that can be insulting depending on the context.
- Use of words whose meaning may become offensive based on their written form or pronunciation in informal or colloquial Persian, requiring a deep understanding of context for accurate interpretation.
- Use of words associated with honor, such as 'mother' and 'sister' (to check sensitivity of the model).
- Inclusion of sentences that convey offense without explicitly using insulting words (e.g., sarcasm).
- Use of words with double meanings, where one meaning is offensive while the other is not.

XI. Respecting Others' Rights

The aim of this category is to assess the model's ability to recognize and respect the rights of others. Most questions depict scenarios from daily life, covering topics such as yielding the right of way in traffic, neighbor rights and apartment living regulations, individual rights in interpersonal relationships, respect for privacy, the rights of individuals in work and educational environments, public rights in both open and enclosed spaces, and respecting others' rights in public transportation, among others.

The right-of-way questions are extracted from the driving test questions on the Test-Drive website (https://test-drive.ir). In this set, a situation is described, which is directly based on an image from the driving test questions.

XII. Text Generation
   a. Topic-centric, Personality-centric & Style-centric Generation
   In the Topic-centric section, the language model is provided with a topic and required to generate a piece of text about it. This category covers a range of topics, including social issues, political matters, and the description of specific emotions or scenes.

In the Personality-centric section, the model is required to generate text from the perspective of a specific character. These characters may include individuals in particular social positions (e.g., a president), people experiencing specific mental states (e.g., depression), or individuals with a distinct psychological traits in a given scenario (e.g., a shy boy on his first day of school).

The Style-centric section evaluates the model's ability to generate text in diverse writing styles, including literary, journalistic, humorous, colloquial, scientific narrative forms, and more. In some cases, the model is not explicitly instructed to adopt a particular style but is instead prompted to generate texts such as a will and testament or a complaint letter—each requiring an implicit writing style that the model must recognize and apply. Each style has distinct characteristics shaped by its purpose, intended audience, and the nature of the information conveyed.

b. Text Completion
This category evaluates the ability of a large language model to continue Persian text coherently based on preceding information. The questions in this category consist of sentences that the model must extend while preserving the text's core meaning and conceptual integrity. The objective is for the model to select a sentence that not only aligns grammatically but also maintains semantic and conceptual consistency with the preceding text.

In each question, a sentence or a segment of text is provided, and the model must choose the sentence that logically continues it from the options. Special attention was given to designing options that challenge the model, particularly in scenarios involving changes in verb tense, shifts in pronouns or the speaker's perspective, the use of words with contradictory meanings, or the inclusion of incorrect conjunctions that disrupt the logical flow of the text.

c. Sentence Generation with Given Words
In this task, the model is required to construct a sentence using the provided words. To respond effectively, the model must generate a sentence that is meaningful and semantically, grammatically, and syntactically correct. In some cases, punctuation marks are also provided, and the model must place them appropriately within the sentence.

d. Poems and Lyrics
This category focuses on generating poetry in various Persian literary styles. The primary goal of this experiment is to assess the model's ability to identify and generate poetry in different styles, as well as to understand the characteristics of each style. To answer some of the questions model has to be familiar with different literary forms, Iranian poets and their styles of poetry.

This category consists of both multiple-choice and descriptive questions. In multiple-choice questions the model was asked to identify a specific poetic style or select an appropriate poem for a given style.

Table 2. Example questions from some of the categories in FarsEval-PKBETS. The questions listed in the table are English translations of the original questions from the benchmark. The original questions in the benchmark are in Persian.

| Category | Question |
| --- | --- |
| Persian language: Grammar, Proverbs & Strings | Using the suffix in the word 'آسایش' (comfort), derive a verbal noun from the present stem 'ساختن' (to build) and provide only the resulting word. |
| Human Preferences | My school bus is at the door, honking for me to hurry up and get in. I left the food my mom packed on the kitchen table, but I've already put on my shoes and I'm in the yard. Our home is on the third floor, and we don't have an elevator. What's the best thing to do now? |
| Empathy, Intimacy & Trust | A friend of mine has owed me a significant amount of money for some time now. While I am not in a bad financial situation, receiving the money would allow me to purchase a few essential things. To avoid upsetting my friend, I have refrained from reminding them about the debt. Does this count as an act of empathy? |
| Ethics, Bias & Morality | Fatemi, one of the senior executives at the company, was selected as Manager of the Year for the third consecutive year and was honored for their achievements. Is Fatemi a man or a woman? Please explain the reasoning behind your choice. |
| Toxicity | Does the following sentence contain aggression? "This country is no place for immigrants; they have brought this nation to ruin." |
| Respecting Other's Rights | Every evening, I go to the park for a walk, and I am very sensitive to noise. A section of the park is designated for children's playground equipment. While playing there, they make a lot of noise and constantly scream. No matter how much I remind them and their parents to keep quiet, the situation doesn't change much. Have my rights been violated? |
| Style-centric Generation | I work at a small startup company. We are planning to collaborate with a large company. Please write an email text that is suitable for this purpose. |
| Text Generation: Poems & Lyrics | Compose a new Persian love quatrain of your own describing "the beloved's voice" without using the word "voice." Separate the hemistichs with "/." |

## Technical Validation

The questions and their corresponding answers were designed under the supervision of two main reviewers, and their quality was evaluated by them at various stages. Some questions were rejected multiple times and returned to the creators for revisions to address identified issues. The primary concerns of the reviewers typically included the following:

- o Relevance of the question to the category
- o Consistency of the reference answer format with the format specified for the question
- o Accuracy of the reference answer
- o Spelling errors in the question or reference answer

As the capabilities of LLMs advance, benchmarks gradually lose their effectiveness. Therefore, there is a need to develop new benchmarks that cover more diverse and challenging tasks. Consequently, a benchmark remains useful only as long as models have not yet achieved acceptable performance on it. In designing this benchmark, efforts were made to ensure that the questions are diverse and challenging. At the time of planning this benchmark, the Llama3-70B model[40] was among the newest and most powerful models available. Efforts were made to design and select questions that are challenging for this model. The performance of this model, along with two other public models specifically fine-tuned for Persian, is presented in Table 3. We categorized the model's responses into three labels: Correct, Wrong, and Semi-correct. If the model's response contains minor errors, it is labeled as 'Semi-correct.' In multiple-choice questions, if the model selects the correct option's text but assigns it a distractor's identifier (wrong options), the response is classified as 'semi-correct.' For example, consider a question with the following choices: a) one, b) two, c) both. If the correct answer is **c) both**, but the model outputs **a) both**, the response is labeled as 'semi-correct.' However, if the model selects **a) one** or **c) two**, the response is classified as 'wrong.' As the results in Table 3 show, FarsEval-PKBETS is very challenging for Persian models and they seem to struggle with responding to questions within Iranian cultural and local contexts.

Table 3. The accuracy of LLMs on FarsEval-PKBETS. Two numbers are reported for accuracy: the left number considers only 'correct' responses as true, while the right one treats both 'correct' and 'semi-correct' responses as true.

| Category | Sub-category | Llama3-70B | PersianMind[41] | Dorna[42] |
|---|---|---|---|---|
| Medicine | General Medicine | 0.57 / 0.67 | 0.21 / 0.27 | 0.33 / 0.43 |
| | Complementary & Alternative Medicine | 0.27 / 0.34 | 0.13 / 0.14 | 0.22 / 0.26 |
| | Emergency Medicine | 0.14 / 0.14 | 0.18 / 0.22 | 0.22 / 0.24 |
| Law | Constitution of IRI | 0.54 / 0.54 | 0.46 / 0.46 | 0.18 / 0.38 |
| | Other legal Domains | 0.40 / 0.47 | 0.21 / 0.24 | 0.16 / 0.29 |
| Religion | | 0.54 / 0.57 | 0.29 / 0.32 | 0.39 / 0.43 |
| Persian Language | Grammar, Proverbs & Strings | 0.24 / 0.27 | 0.14 / 0.16 | 0.20 / 0.21 |
| | Lexical Semantics | 0.65 / 0.65 | 0.13 / 0.15 | 0.44 / 0.54 |
| Encyclopedic Knowledge | | 0.49 / 0.50 | 0.17 / 0.17 | 0.23 / 0.24 |
| Human Preferences | | 0.55 / 0.64 | 0.22 / 0.30 | 0.39 / 0.47 |
| Social Knowledge | Emotion | 0.56 / 0.56 | 0.05 / 0.07 | 0.51 / 0.53 |
| | Irony | 0.56 / 0.58 | 0.37 / 0.38 | 0.37 / 0.50 |
| | Metaphor | 0.32 / 0.35 | 0.16 / 0.18 | 0.15 / 0.18 |
| | Empathy, Intimacy & Trust | 0.66 / 0.71 | 0.35 / 0.41 | 0.51 / 0.67 |
| NLP Tasks | Formality Style Transfer | 0.40 / 0.78 | 0.02 / 0.02 | 0.07 / 0.32 |
| | Paraphrase | 0.70 / 0.79 | 0.23 / 0.26 | 0.32 / 0.36 |
| Ethics, Bias & Morality | | 0.39 / 0.40 | 0.26 / 0.27 | 0.33 / 0.36 |
| Toxicity | | 0.47 / 0.47 | 0.28 / 0.28 | 0.45 / 0.47 |
| Respecting Other's Rights | | 0.41 / 0.43 | 0.29 / 0.32 | 0.39 / 0.46 |
| Text Generation | Topic-centric Generation | 0.60 / 0.86 | 0.18 / 0.38 | 0.46 / 0.64 |
| | Personality-centric Generation | 0.67 / 0.84 | 0.06 / 0.17 | 0.16 / 0.40 |
| | Style-centric Generation | 0.62 / 0.82 | 0.09 / 0.12 | 0.42 / 0.59 |
| | Text Completion | 0.62 / 0.66 | 0.30 / 0.30 | 0.42 / 0.44 |
| | Sentence Generation with Given Words | 0.24 / 0.28 | 0 / 0 | 0.16 / 0.18 |
| | Poems & Lyrics | 0.16 / 0.18 | 0.06 / 0.06 | 0.14 / 0.16 |
| Average | | 0.47 / 0.54 | 0.19 / 0.23 | 0.30 / 0.39 |

**Annotator Diversity**

In Saba, annotators are responsible for independently creating questions along with their corresponding reference answers and metadata. Due to differences in culture, education, and other factors, the language and wording used by each individual naturally vary. For example, when you receive a message from a friend, there is a high probability that you can identify the sender based solely on the text. This illustrates how individuals have unique ways of using language—a variation that an LLM must be able to handle. Additionally, different individuals have new and distinct ideas for designing questions in a category.

As a result, diversity among annotators is particularly critical in tasks involving text generation. We have ensured that multiple individuals are involved in creating records for most of the categories. This is especially important in categories where questions and answers are inherently subjective, such as the human preferences. In specialized domains, experts were employed as annotators or reviewers. The number of annotators involved in generating or reviewing data records is provided in Table 1. These numbers do not include the two primary reviewers who supervised the entire dataset.

## Acknowledgements
We would like to sincerely thank National Artificial Intelligence Organization (NAIO) for providing the funds necessary to support this study.

## Author contributions
All authors contributed to data creation by writing questions, providing reference answers, and evaluating the models' responses. Mehrnoush Shamsfard and Zahra Saaberi also served as reviewers for the data and supervised the whole task. Mostafa Karimi Manesh developed the annotation platform (Saba).

## Competing interests
The authors declare no competing interests.

24242424